\pdfoutput=1

\RequirePackage[T1]{fontenc}

\PassOptionsToPackage{dvipsnames}{xcolor}
\documentclass[letterpaper, 10 pt, conference]{ieeeconf}
\IEEEoverridecommandlockouts
\usepackage{iftex}
\ifPDFTeX
  \usepackage[utf8]{inputenc}
\fi

\usepackage{amsmath}
\usepackage{amssymb}
\usepackage{bm}

\usepackage{graphicx}
\graphicspath{{figures/}}
\usepackage{booktabs}     
\usepackage{multirow}
\usepackage{subcaption}

\usepackage[dvipsnames]{xcolor}
\usepackage{siunitx}
\usepackage{cite}

\usepackage{url}
\usepackage[hidelinks, breaklinks=true]{hyperref}

\newcommand{\Dbar}[1]{\bar{D}_{#1}}
\newcommand{\eps}{\varepsilon}
\newcommand{\ts}{\mathrm{ts}}

\newcommand{\figtaskref}[1]{Fig.~\ref{fig:task#1}}

\begin{document}
\bstctlcite{IEEEexample:BSTcontrol}

\typeout{PAGE-GEOMETRY: textwidth=\the\textwidth\space columnwidth=\the\columnwidth\space textheight=\the\textheight}

\title{\LARGE \bf Same Action, Different Outcome: \\Variability in Dynamic Cloth Manipulation}
\author{Mahed Dadgostar$^{1}$, Guillem Alenyà$^{1}$ and Júlia Borràs$^{1}$
\thanks{$^{1}$
The authors are with Institut de Robòtica i Informàtica Industrial, CSIC-UPC, Spain;
This work has been funded by project PID2023-152259OB-I00 (Chloe-MAP) funded by MCIU/ AEI /10.13039/501100011033 and by ERDF, UE and project HORIZON-WIDERA-2023-ACCESS-02-01-101159522 (Romandic). 
Contact: {\tt\small \{mdadgostar, galenya,jborras\}@iri.upc.edu}. 
Claude Code (Fable 5.1) was used to assist with coding, including code for figure generation. All ideas and final content are entirely our own.
}
\thanks{Project webpage: \url{https://cloth-var.github.io} }
}

\maketitle

\begin{abstract}
Although cloth is known to exhibit different outcomes under repeated fast dynamic motions, even when the same trajectory is applied, this variability has not yet been systematically characterized. Quantifying it is essential to assess the reliability of learned manipulation policies and the extent to which simulation can reproduce real-world behavior.
To study this, we execute the same trajectory ten times across four dynamic tasks, two of which are novel, each tested with three cloths of very different properties and at up to three execution speeds, with a total of 269 recorded rollouts. For all of them, we record small marker positions on the cloth and synchronized stereo camera. We then formalize different metrics to quantify variability, 
and our results show how it is significant in every test condition, 
in most cases one to three orders of magnitude above the repeatability inherent to the robot and sensing noise.
Our results also show variability is driven mainly by the cloth physical properties but also grows with speed.
By replaying all the rollouts in four calibrated modern cloth simulators, we show that none of them can reproduce the variability magnitude we observed in real cloth, nor its ordering. 
Together with our analysis, we publish the dataset with synchronized OptiTrack, vision and robot logs and their corresponding simulator twins.
\vspace{-3mm}
\end{abstract}

\section{Introduction}
\label{sec:intro}

Most available deformable object manipulation datasets contain slow folding tasks or unfolding pick-and-place, 
in which inertial effects can be ignored, and the motion is referred to as quasi-static manipulation \cite{mason1993dynamic, mason2001mechanics}. In contrast, dynamic manipulation consists of motions in which the inertia effects and accelerations of the target objects are significant during execution: a bimanual setup can fold a cloth slowly by grasping its two corners, but a single robotic arm can perform the same task by leveraging inertia to handle it faster \cite{jangir2020dynamic, hietala2022visual}.

With advances in foundational models for natural language processing and computer vision, researchers have extended this paradigm to robotics through vision-language-action (VLA) models \cite{black2024pi0, kim2024openvla}. However, the cloth manipulation tasks they perform are typically quasi-static, such as folding laundry \cite{black2024pi0}, and their perception and inference run at a few hertz \cite{kim2024openvla}. In the dynamic manipulation regime, aerodynamic, frictional and other environmental effects affect the reliability and success of the task. The works that do manipulate cloth dynamically \cite{ha2021flingbot, canberk2023clothfunnels, hietala2022visual} report only policy success rates, which  do not characterize the variability in cloth behavior that arises even across nominally identical trials.

\begin{figure}[t]
    \centering
    \includegraphics[width=0.9\linewidth]{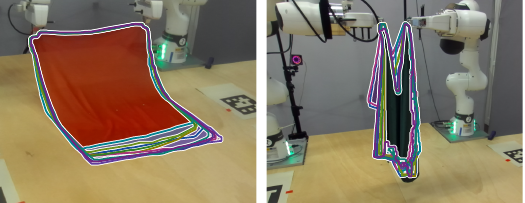}
    \caption{\small Terminal cloth outlines for ten identical policy rollouts, shown on one frame, for 2 tasks: Swing to place (right) and gripper inward motion (left).}
    \label{fig:teaser}
    \vspace{-4mm}
\end{figure}

Another challenge, often overlooked in sim-to-real transfer, is the inconsistent cloth behavior between the real environment and the simulator, especially in dynamic tasks. For example, if one executes a swinging trajectory in a deterministic physics simulator for $N$ samples, the cloth states across all samples remain the same if the cloth's initial state and grasp have not changed. However, if the same experiment is performed in the real world, the cloth-state trajectories diverge across rollouts (Fig.~\ref{fig:teaser}). This is because the available simulators account for aerodynamics only coarsely \cite{coltraro2025aerodynamic} and not at all for air turbulence or micro-level changes in the cloth state, such as the grasp posture or the cloth history of wrinkles and fold lines. This means that this issue cannot be addressed even with domain randomization \cite{tobin2017domain, matas2018simtoreal}, since these effects are not parameters of the simulator. Modeling these aerodynamic effects can be extremely complex; in this sense, one could say that dynamic manipulation of cloths is nondeterministic. However, like aleatoric uncertainty \cite{huellermeier2021aleatoric}, it can be characterized by quantifying how much variability remains after accounting for everything the robot can observe and control. This work aims to quantify precisely this residual variability.

To do so, we replay the same commanded trajectory ten times with the same trial conditions. We evaluate across four different tasks, each performed with three cloths, and up to three execution speeds to show that this variability exists and to characterize its magnitude, controllable factors, and modes. We believe that improving the modeling parameters of cloth dynamics can only go so far, and that identifying the controllable factors of this interaction inconsistency tells us how to mitigate it. For instance by acting more slowly on a cloth sensitive to aerodynamic effects, or estimating how much data a policy needs and when gathering more data cannot improve task success. 

The main contributions of this work are:
\begin{itemize}
    \item A formalism to show that the same dynamic manipulation of cloths yields different outcomes depending on the nature of the manipulation and the cloth's features, together with the analysis of the results on four dynamic tasks. 
    \item An evaluation of four cloth simulators, calibrated per cloth and replaying the measured inputs, showing that this variability is poorly modeled: no engine reproduces its magnitude or its ordering reliably.
    \item A dataset of 269 motion-captured rollouts, with synchronized stereo video, robot logs, and corresponding simulator twins in 4 different simulators, for 4 dynamic cloth manipulation tasks.
\end{itemize}

The dataset, available on the paper's website\footnote{\url{https://cloth-var.github.io}}, contains two of the most commonly studied dynamic tasks: swing-to-place flat, also known as fling \cite{ha2021flingbot, canberk2023clothfunnels}, and dynamic folding on the table, using a single hand \cite{hannus2024dynamic, hietala2022visual,caldarelli2026koopman}.  It also includes two novel dynamic tasks that, to the best of our knowledge, have not been studied: moving the grippers inward toward each other, with and without a swing  (Fig.\ref{fig:teaser} - right). Humans perform this motion to fold in the air.
\thispagestyle{empty}
\pagestyle{empty}

\section{State of the Art}
\label{sec:sota}

Several examples exist in the literature that prove the advantage of dynamic manipulation. For instance, quasi-static pipelines smooth garments by sequences of pick-and-place actions \cite{seita2020deep, ganapathi2021learning}. A fling unfolds a cloth in a fraction of those interactions \cite{ha2021flingbot}, and other dynamic actions followed: flings that align garments \cite{canberk2023clothfunnels}, single-arm folding \cite{jangir2020dynamic, hietala2022visual, caldarelli2026koopman} and placing \cite{mavsar2026dynamic}.

That nominally identical dynamic executions leave the cloth in different states has been observed but not characterized. Coltraro et al.~\cite{coltraro2025tracking} repeat one dynamic laying motion fourteen times on their lightest fabric and find a considerable spread of the final corner positions. Hietala et al.~\cite{hietala2022visual} record ten open-loop dynamic folds per fabric, but do not discuss their spread. Neither work, nor any other we know of, has analyzed the spread in relation to the fabric, the speed or the grasp, or separated it from the repeatability of the robot. 
However, in the context of flexible sheets, similar studies have proven useful showing how in a flow these sheets flap above a critical speed and become chaotic at higher speeds \cite{shelley2011flapping}, so micro-differences between trials are amplified in flight.

In works where the variability is handled, it is assumed rather than measured; robust MPC certifies constraint satisfaction against a bounded disturbance set \cite{li2026cordsls}, and iterative learning control assumes the disturbance repeats from trial to trial \cite{bristow2006survey}, as does the trial-to-trial refinement of deformable-object actions \cite{chi2022irp, suresh2026taskilc}. Which of these assumptions dynamic cloth manipulation satisfies is unknown, because the distribution of its outcome has not been reported; measuring it requires the cloth state in flight, where self-occlusion and motion blur defeat vision \cite{longhini2025review}. To overcome this challenge, we use motion capture as in \cite{coltraro2025tracking}, manually linking markers that go occluded for short periods and in addition, we add a registered stereo camera. In \cite{coltraro2025tracking}, they prove that these small markers are light enough not to affect the dynamics of the cloth. Therefore, our dataset improves on \cite{coltraro2025tracking} by adding novel dynamic tasks, including synchronized image perception.

\section{Problem Formulation}
\label{sec:formulation}

\subsection{Interaction variability}
\label{sec:problem}
Interaction variability is the non-repeatability of the cloth state under the same trajectory and a nominally identical initial state. Let $\pi$ be the manipulation policy and $y_t$, $r_t$, $u_t$ the observed cloth state, robot state, and control input at time $t$. Here $\pi$ replays an open-loop trajectory, so $u_t = \pi(y_t, r_t)$ is deterministic. The next cloth state is

\begin{equation*}
  y_{t+1} \sim p_{\mathrm{real}}(y_{t+1} \mid y_t, r_t, u_t),
\end{equation*}
where $p_{\mathrm{real}}$ is the probability distribution over the next state given the current
observation and control input.

We define the \emph{configuration} $c \in \mathcal{C}$ of a robot rollout as the task policy, object used and velocity of the policy execution. In addition, let the rollout $i$ be defined as $y^{(i)}_{0:T} := \{y^{(i)}_t\}_{t=0}^{T}$, with $T$ the trajectory horizon. Then, we define a rollout feature $z$ as
\begin{equation}
  z^{(i)} := \phi\bigl(y^{(i)}_{0:T}\bigr), \quad
  z^{(i)} \sim p_{\mathrm{real}}(z \mid \pi), \; i = 1, \ldots, N_c,
  \label{eq:rollout}
\end{equation}
where $\phi(\cdot)$ extracts the task-dependent feature, $N_c$ is the number of rollouts for a given configuration and $p_{real}$ represents the probability distribution of $z$ over repeated samples of the same policy.

However,  every real-world experiment has some unavoidable noise in execution inherent to the robot repeatability or sensing method, which we call floor variability $\eps$. 
We measure the variability of a feature $z$ as the expected distance between different rollouts, and how different it is from the floor value $\eps$. That is,
\begin{equation}
  \mathbb{E}\bigl[ d\bigl(z^{(i)}, z^{(j)}\bigr) \bigr] > \eps, \qquad \forall i,j, \text{ } i \neq j,
  \label{eq:criterion}
\end{equation}
where $d$ is a measure of task-dependent distance between different outcomes.

\newlength{\panela}
\newlength{\panelb}
\setlength{\panela}{0.257\textwidth}
\setlength{\panelb}{0.321\textwidth}

\begin{figure}
    \centering
    \includegraphics[width=\linewidth]{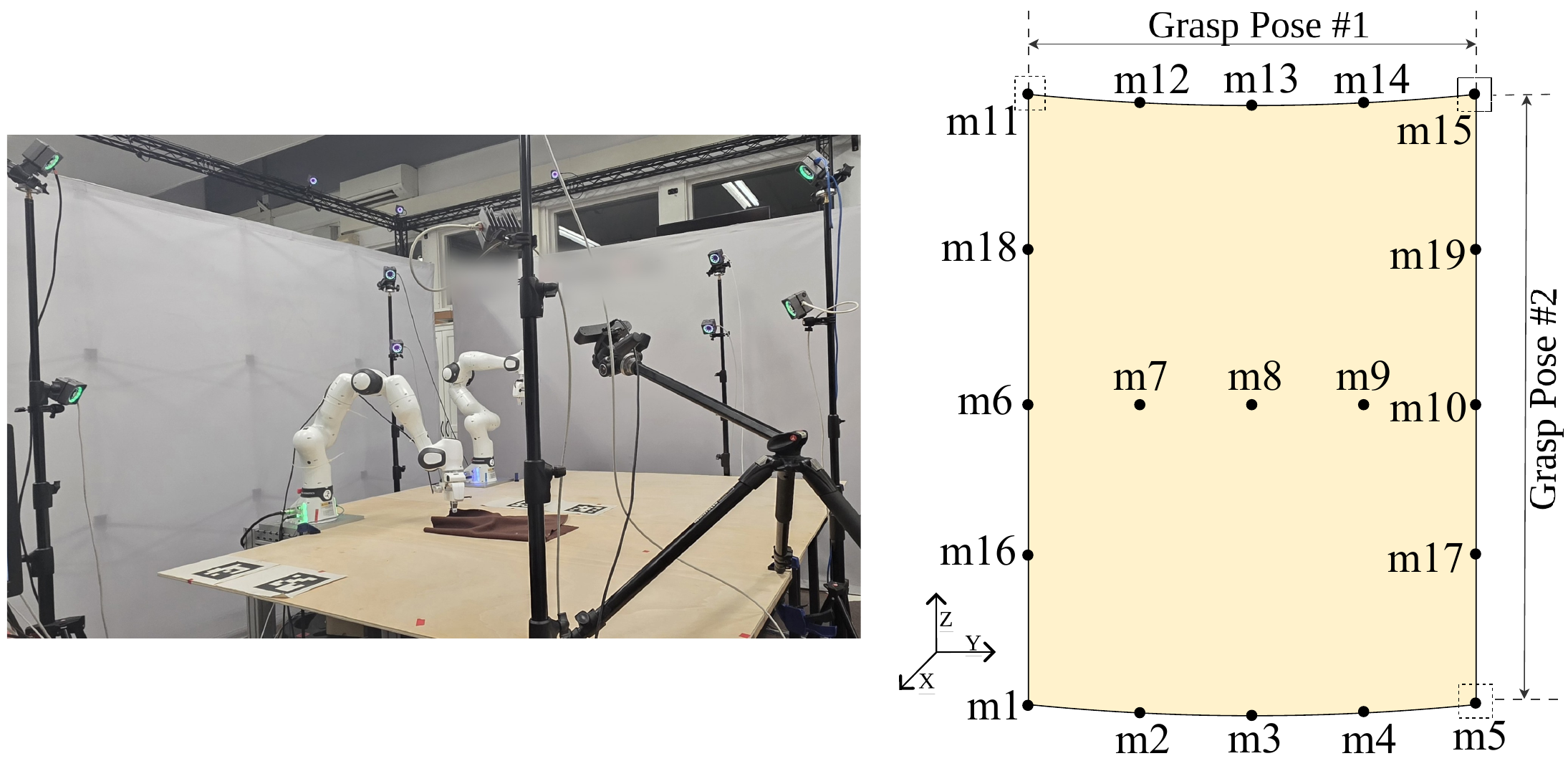}
    \caption{(Left) Bimanual FR3 cell with OptiTrack cameras to
           track retroreflective markers on the cloth while the two arms execute
           the grasp poses. (Right) Marker distribution.}
    \label{fig:setup-all}
\end{figure}

\begin{table}[tb]
  \centering
  \caption{Fabric properties following\cite{garciacamacho2024standardization}}
  \label{tab:fabric}
  \footnotesize
  \setlength{\tabcolsep}{5pt}
  \begin{tabular}{@{}lrrr@{}}
    \toprule
                                                & brown  wool     & green satin     & red chiffon          \\
    \midrule
    mass (\si{\gram})                           & 139        & 35         & 27           \\
    areal density (\si{\gram\per\square\metre}) & 400        & 100        & 76           \\
    friction coefficient $\mu$                  & 0.68       & 0.48       & 0.40         \\
    elasticity$^\dagger$ (\si{\percent})     & 8 / 1 / 17 & 1 / 12 / 18 & 20 / 38 / 13 \\
    drape stiffness (\si{\percent}) & 9.5        & 5.9        & 3.1          \\
    \bottomrule
  \end{tabular} \\
  \vspace{1mm}
  $^\dagger$ Elasticity along the short, long and diagonal edges, respectively.
  \vspace{-3mm}
\end{table}

In our experiments, the cloth state $y_t$ at time $t$ and rollout $i$ is defined by the tracked marker positions
$p^{(i,m)}_t \in \mathbb{R}^3$ for each cloth marker $m \in \mathcal{M}$. One of the relevant features to study variability is the
location of each marker at the end of the execution, that is, $q^{(i,m)}$. We define the distance between rollouts $i$ and $j$ as the mean distance between corresponding terminal
points of all markers: 
\begin{equation}
  D_c(i, j) := \frac{1}{|\mathcal{M}_{ij}|}\sum_{m \in \mathcal{M}_{ij}}
  \bigl\lVert q^{(i,m)} - q^{(j,m)} \bigr\rVert_2 ,
    \label{eq:Dcij}
\end{equation}
where $\mathcal{M}_{ij}$ contains only the markers observed in both rollouts. Marker identity is known because we manually supervised the labels of all visible markers, so no matching is needed.
The mean over all rollout pairs, $\Dbar{c}$ represents the dispersion of the task configuration:
 \begin{equation}
  \Dbar{c} := \binom{N_c}{2}^{-1} \sum_{i < j} D_c(i, j).
  \label{eq:dbar}
\end{equation}
This formula  estimates $\mathbb{E}[d]$  in Eq.~\eqref{eq:criterion}.

Per marker, the sample covariance of $\{q^{(i,m)}\}_i$ is drawn as a
\SI{95}{\percent} ellipse ($\chi^2$, two degrees of freedom) in the figures; a non-degenerate ellipse shows $\operatorname{Var}[z \mid \pi] \succ 0$ for that marker.

To estimate the floor variability $\eps$, we note that the robot is highly repeatable, with the logged tool pose and the markers on the grippers repeating to less than \SI{0.4}{\milli\metre} in every rollout. Instead, we use the cloth markers at the grasped corners, next to the gripper fingers. By applying the same measures to these markers, as they move with the robot, they measure the variability inherent to the robot, the grasp, but also the sensing technology. That is, for a given task configuration $c$, the floor variability is
\begin{equation}
  \eps_c := \binom{N_c}{2}^{-1} \sum_{i < j} \frac{1}{|\mathcal{H}_{ij}|}
  \sum_{m \in \mathcal{H}_{ij}} \bigl\lVert q^{(i,m)} - q^{(j,m)} \bigr\rVert_2 
  \label{eq:floor}
\end{equation}
where $\mathcal{H}$ is the set of markers at the grasped corners.
The configuration is non-repeatable when $\Dbar{c} > \eps_c$. This is the empirical form of Eq.~\eqref{eq:criterion}.

The speed contrast of a cloth is the ratio of its fastest to its slowest
dispersion without regrasp,
\begin{equation}
  \Gamma_D := \Dbar{c(\text{cloth},\ \min \ts)} \,\big/\,
  \Dbar{c(\text{cloth},\ \max \ts)} ,
  \label{eq:gamma}
\end{equation}
where $\ts$ multiplies the trajectory duration, so $\min \ts$ is the fastest
execution of that cloth. $\Gamma_D > 1$ means the faster execution scatters
more.

\section{Experimental protocol}
\label{sec:setup}

\subsection{Experiment Setup}
\label{sec:hardware}
Two Franka Research 3 arms with default Franka Hands are placed side by side across a table (Fig. \ref{fig:setup-all}-left). 

An OptiTrack motion-capture system tracks markers at \SI{100}{\hertz} in a world frame with $x$ away from the robot base and $z$ up. A stereo camera (ZED~2i) records the bimanual-with-cloth scene at \SI{30}{\hertz}. Additionally, four ArUco tags, each with tracking markers placed on the table top-right corner, are placed on the table to provide the transformation matrix between the OptiTrack origin and the camera. Each Franka also has three markers on its end effector. 
Cloths' corners are grasped so that their markers are visible throughout the episode.

We selected three cloth objects with the size of $0.7 \times \SI{0.5}{\metre}$. Each cloth has 19 markers shown in Fig.~\ref{fig:setup-all}. Depending on the task, we used different grasp locations, either grasping only one corner, the short edge (grasping markers m11 and m15) or the long edge (markers m5 and m15). The cloths used have the same size but extremely different physical cloth properties, which we measured following~\cite{garciacamacho2024standardization}, with the results in Table~\ref{tab:fabric}. The brown cloth is thick wool fabric with high stiffness and low elasticity. The green satin is much lighter with less stiffness and friction but a similar elasticity. Finally, the red is a very light chiffon fabric that is very elastic, very soft, and has the lowest friction coefficient.
\begin{figure*}[t]
  \centering
  \includegraphics[width=\textwidth]{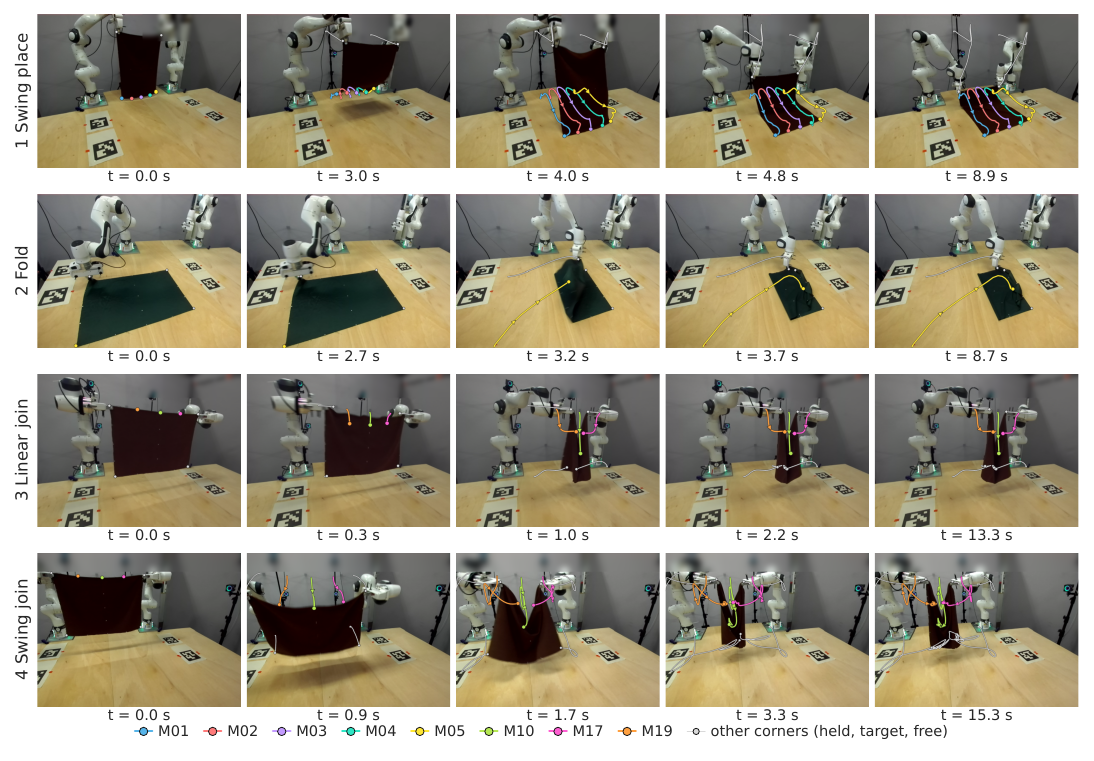}
  \caption{Each row shows 5 frames of the recorded tasks. The markers used to study the variability for each task are colored and tracked in the camera view.
  Other relevant markers are in gray.}
  \label{fig:tasks}
  \vspace{-3mm}
\end{figure*}

\subsection{Protocol}
\label{sec:protocol}
Table \ref{tab:dataset} summarizes all the combinations of tasks, speed, regrasps, and cloth fabrics; each of the 27 configurations includes ten identical rollouts\footnote{ Except for task~4 with brown cloth, $\ts = 1.5$, with only 9 recorded}. We have two types of configurations:
\begin{enumerate}
  \item Fixed grasp: The grasp posture for all repetitions of the configuration remains the same. The bimanual system starts each rollout from the same initial pose.
  \item Regrasp: To study the effect of regrasping, we manually regrasp the cloth at each rollout. Then, the robot starts each rollout at the same initial pose. 
\end{enumerate}

Three factors are varied: (i) the object, (ii) motion speed, (iii) the grasp for only a few configurations.
To vary speed, given a policy, we change the time scale of its execution by a factor $\ts$ relative to the speed at which it was recorded, with $\ts < 1$ meaning the episode was executed faster than recorded and $\ts > 1$ for slower execution.

The trajectories for tasks~1, 2, and~4 are collected through a teleoperated demonstration in an open-loop manner, while a linear scripted trajectory planned in joint-space is used for task~3. The teleoperation device was a rigid body with OptiTrack markers, tracked in 6-D.

\begin{table}[!tb]
  \centering
  \caption{The four tasks and their 27 configurations.}
  \label{tab:dataset}
  \footnotesize
  \setlength{\tabcolsep}{5pt}
  \begin{tabular}{@{}rllll@{}}
    \toprule
    \# & Task & Object$^{\dagger}$ & Timescale $\ts$ & Regrasp\\
    \midrule
    1 & Swing place   & 1 cloth (b)  & 0.8, 1, 1.2 & No \\
     &                & 2 cloths (g, r) & 0.65, 1, 1.35 & No\\
    2 & Fold         & 3 cloths & 1, 1.4  & No\\
    3 & Linear join   & 3 cloths & 1 & No\\
      &               & 1 cloth (g) & 1 & Yes\\
    4 & Swing join    & 3 cloths & 1, 1.5 & No \\
      &               & 2 cloths (g, r) & 1      & Yes\\
    \bottomrule
  \end{tabular}\\
  \vspace{1mm}
    $^{\dagger}$ b, g and r stand for brown, green and red cloths, respectively
   \vspace{-3mm}
\end{table}

\subsection{Recorded Tasks}
\label{sec:tasks}
Fig.~\ref{fig:tasks} summarizes the collected tasks showing one rollout of each. 

\smallskip\noindent\textbf{Task 1: swing place.}
Both arms hold the short edge m11--m15, swing the cloth forward and lay it
flat without releasing it (see Fig.~\ref{fig:tasks}- row 1). The feature measured is where the free edge m1--m5 (bottom edge) lands and rests. The resting pose is defined as the mean pose for the last 50 frames without motion. 
 Therefore, for this task, $\Dbar{c}$ of Eq.~\eqref{eq:dbar} is evaluated on the resting pose of
 m1, $\dots $, m5.  To compute distances, we project all coordinates on the table (we use only $x$ and $y$).
Finally, the floor variability $\eps_c$ is obtained evaluating  Eq.~\eqref{eq:floor} for the grasped corners m11 and  m15.

\smallskip\noindent\textbf{Task 2: dynamic fold.}
The right arm holds the corner m1 and folds it towards m11 (see Fig.~\ref{fig:tasks}-row 2). The edge m11--m15 stays in place and the free corner m5 is aimed to land on the corner m15.
In this case, $\Dbar{c}$ is obtained by evaluating Eq.~\eqref{eq:dbar} for only the resting pose of the marker m5, taking also only 2-dim distances on the table as before, and $\eps_c$ is measured on the held corner m1.

\begin{figure*}[t]
  \centering
  \includegraphics[width=\textwidth]{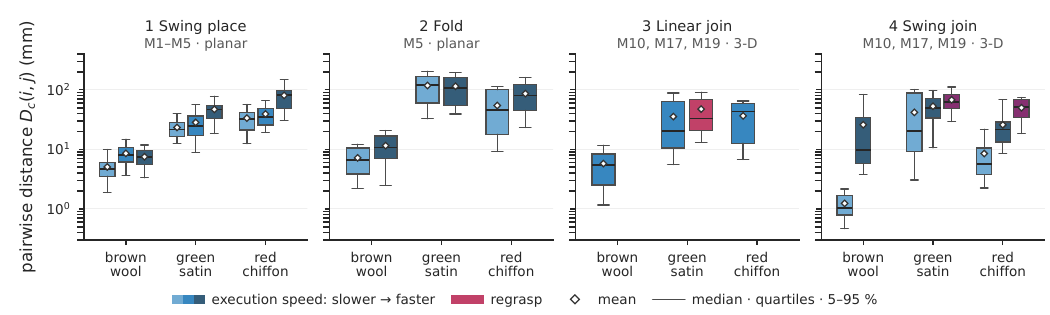}
  \caption{ We show the variability amount for all tasks and their configurations with the whisker boxes of all the pair distances $D_c(i,j)$ in all rollouts per configuration. The variability is orders of magnitude different depending on the task; that is why the y-axis is on a logarithmic scale.}
  \label{fig:cross}
  \vspace{-2mm}
\end{figure*}

\smallskip\noindent\textbf{Task 3: linear inward gripper motion (linear join).}
The grippers hold the corners m5 and m15 and move simultaneously along a straight line toward each other, stopping before contact. The cloth ends hanging crumpled in the air (see Fig.~\ref{fig:tasks}- row 3). 
For this task we analyze the motion of all markers, with results reported on the paper website for space reasons. Here, we report the outcome of the top edge.
We evaluate $\Dbar{c}$ over the three free edge markers, $\mathcal{M} = \{\text{m17, m10, m19}\}$, while m5 and m15 are held. 
This task is interesting because, if we consider the line m5--m15 where the grippers hold the cloth, after the inward motion the cloth buckles to one of the sides. As with our robot setup the line formed by the grippers is aligned in the y-axis; we can consider the x-coordinates of the free markers m17, m10 and m19 as the buckling distance $d_{\perp}$ between each marker and the line formed by the line m5--m15:
\begin{equation}
  d^{(i,m)}_{\perp} = p_x^{(i,m)}, \text{ for rollout $i$ and marker $m$} \label{eq:dperp}
\end{equation}
In addition, we count the number of rollouts where a marker falls in each of the sides 
\begin{equation}
  f^{+}_{c,m} = \bigl|\{ i : d^{(i,m)}_{\perp} > 0, \forall i \in c \}\bigr|
  \label{eq:line}
\end{equation}
or equivalently $f^{-}_{c,m}$ for the other side, with negative $d_{\perp}$.

\smallskip\noindent\textbf{Task 4: swing inward gripper motion (swing join).}
Task~4 has the same goal as task~3, but introducing dynamics. The arms swing the
cloth out and perform the simultaneous move toward each other, stopping before touching (see Fig.~\ref{fig:tasks} - row 4). We can repeat the exact same measurments than in task 3. The introduction of the dynamic swing in this task is meant to control the direction of the buckling  side. This is a complex dynamic task that, to the best of our knowledge, has not been studied before.

\subsection{Simulator twins} \label{sec:sim-protocol}
To understand how this impact the learning of dynamic cloth manipulation we need to study to what extent this variability is reproduced in existing simulators. We replay every rollout of tasks~1--4 without regrasp in four cloth simulators, each receiving:
\begin{enumerate}
  \item Starting pose, the 19 markers fitted onto a $21 \times 29$ grid by inextensible least squares.
  \item The gripper poses at \SI{100}{\hertz}, pinching one or two corner vertices depending on the tasks, similar to their real-world grasp postures.
  \item The table plane is obtained by taking the mean of the ArUco markers' z-axis which are placed on the table.
\end{enumerate}
 The simulated cloth is relaxed quasi-statically under the pinch, then driven. The engines are  Clothilde~\cite{coltraro2022inextensible,coltraro2025aerodynamic}, the PhysX cloth of Isaac Lab~\cite{mittal2023orbit}, the PBD cloth of Genesis~\cite{genesis2024} and the VBD cloth of Newton~\cite{newton2025}, all with self-collision on.

We calibrate each simulator for each cloth, one set of material parameters for all four tasks.
We minimize the mean distance between simulated and real markers over nine training rollouts per cloth covering every task and speed. We search a coarse grid over the physical range of each engine's stiffness, damping, friction, contact and air-resistance parameters, then refine the best point one parameter at a time. On eight held-out rollouts per cloth, two per task at the medium speed, the marker error is \SIrange{37}{48}{\milli\metre} (Clothilde), \SIrange{70}{78}{\milli\metre} (Isaac Lab), \SIrange{82}{90}{\milli\metre} (Genesis) and \SIrange{79}{94}{\milli\metre} (Newton).

The simulated markers are written back in the motion-capture file format, so the twin dataset runs through the same analysis code as the real one. Then, per configuration, we compare the simulated dispersion with the real one. A full statistical analysis comparing simulation and real is reported on the paper's website.

\begin{figure}[bt]
  \centering
  \includegraphics[width=\columnwidth]{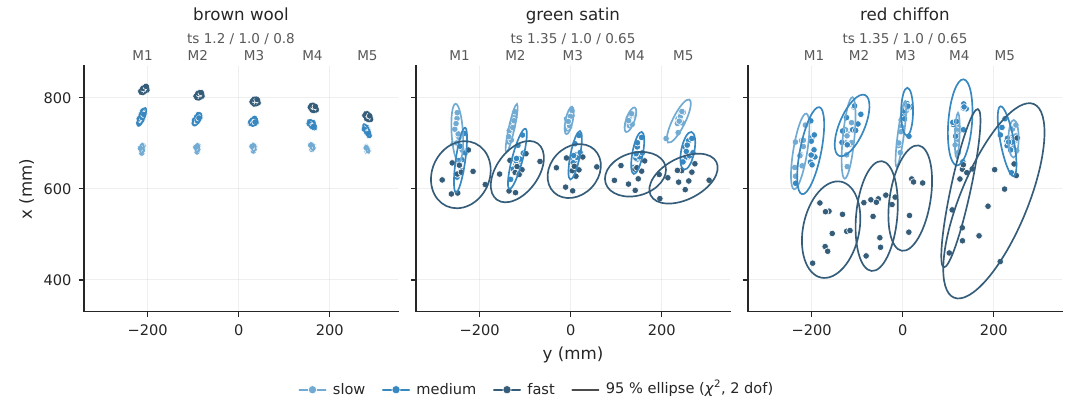}
  \caption{Resting poses of the markers of the bottom edge for the swing place task (Task~1). It shows increasing variability for lighter cloths as well as with higher speeds.}
  \label{fig:task1}
\end{figure}

\begin{figure}[bt]
  \centering
  \includegraphics[width=\columnwidth]{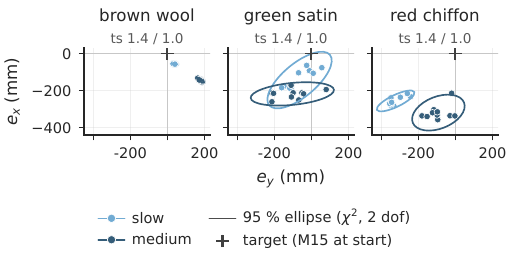}
  \caption{Resting poses of the free marker m5, with respect to the target pose for the task of dynamic folding (task 2). Variability of the lighter cloths are significantly bigger than for the thick brown cloth.}
  \label{fig:task2}
  \vspace{-3mm}
\end{figure}

\begin{figure*}[tb]
  \centering
  \includegraphics[width=0.9\textwidth]{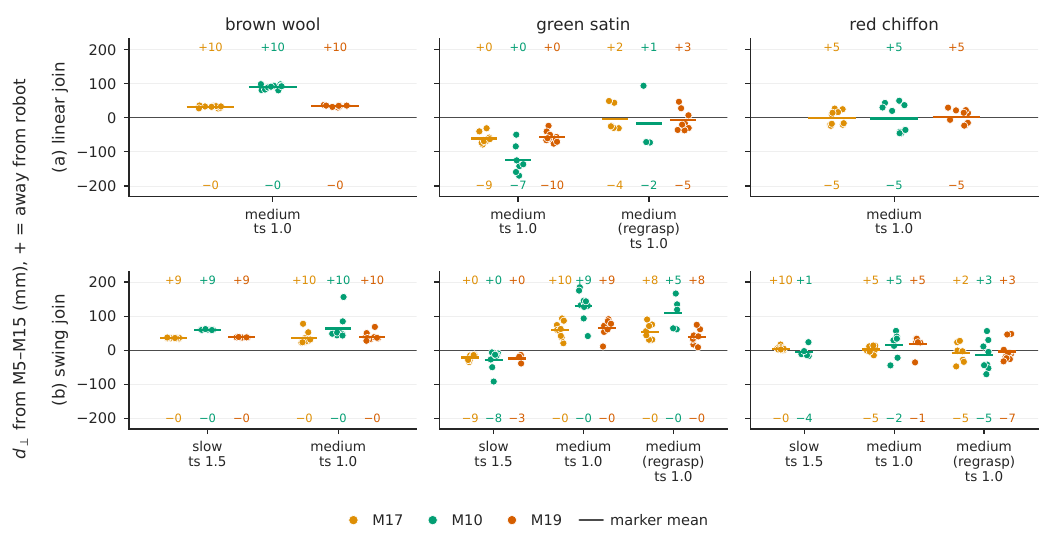}\begin{subfigure}{0pt}\phantomcaption\label{fig:task3}\end{subfigure}\begin{subfigure}{0pt}\phantomcaption\label{fig:task4}\end{subfigure}
  \caption{For tasks 3 and 4, this graphic shows the amount of buckling (distance to line m5--m15 held by the grippers) described in  Eq.~\eqref{eq:dperp} for the visible top cloth edge markers. The numbers at the top and bottom count the buckling side, using Eq.~\eqref{eq:line}. }
  \label{fig:joins}
  \vspace{-3mm}
\end{figure*}
  
\section{Results and Analysis}
\label{sec:results}

\subsection{Comparison of Variability Across Tasks}
\label{sec:res-cross}

We give an overall idea of the variability amount across the different tasks at the different configurations in Fig.~\ref{fig:cross}. The graphic shows the whisker boxes for the pair distances $D_c(i,j)$ of each of the 27 configurations, and their means $\Dbar{c}$ described for each task in the previous section following Eq. \ref{eq:Dcij} and \ref{eq:dbar}. 
The y-axis is logarithmic because  $\Dbar{c}$ spans from 1.2 mm, corresponding to the brown cloth slow swing join task, up to 118 mm for the folding with the green satin, a factor of 96. Every no-regrasp configuration exceeds the floor variability of its grasped markers (see Eq.~\eqref{eq:floor}) by at least $2.4\times$ and most by one to three orders of magnitude. Therefore, it is clear that in all cases, it's a non-negligible phenomenon.  

We observe how the variability changes dramatically, being the cloth type the dominant factor. If we group the 24 no-regrasp configurations, by task or by speed, and measure how much of the spread of $\log_{10}\Dbar{c}$ lies between the groups, the cloth explains 61\,\%, the task 10\,\% and the execution speed only 6\,\% (each factor is computed independently because the set of configurations is not homogeneous, so the shares do not sum to one). The brown cloth, the heaviest and highest-friction fabric, is, as expected, the most repeatable in every task except the fast swing join, where it reaches the same variability as the red cloth. This could be explained because a stiff cloth buckles more once enough dynamics are used. In contrast, the two light cloths scatter three to ten times more.

Regarding speed contrast, defined in Eq. \ref{eq:gamma}, the only one below 1 corresponds to the dynamic fold with the green satin cloth, for which $\Gamma_D=0.97$, meaning that for this task slower motion gets slightly more dispersion than the faster one. This happens in tasks where the optimal speed to achieve the task is high. However, for the rest of the tasks, $\Gamma_D>1$, meaning higher velocity increases variability. The maximum value of $\Gamma_D$ corresponds to the brown wool in the swing inward gripper motion, with $\Gamma_D = 21$. That is because this task is very repeatable with slow speed (only 1.2mm of variability), but with the faster motion the variability increases very fast (to 25.9 mm). In contrast, with lighter cloths, even with slow speeds the variability is substantial, resulting in smaller ratios.

Regrasping the cloth before each rollout raises the floor variability to 12--27\,mm instead of 2--4\,mm. However, the outcome still exceeds this floor by 2.6--4.2$\times$. The corner sits differently in the fingers after a regrasp, and the cloth adds its own variability on top. However, compared to the similar non-regrasped task, we see that regrasping does not significantly affect variability.

\begin{figure*}[t]
  \centering
  \includegraphics[width=0.9\textwidth]{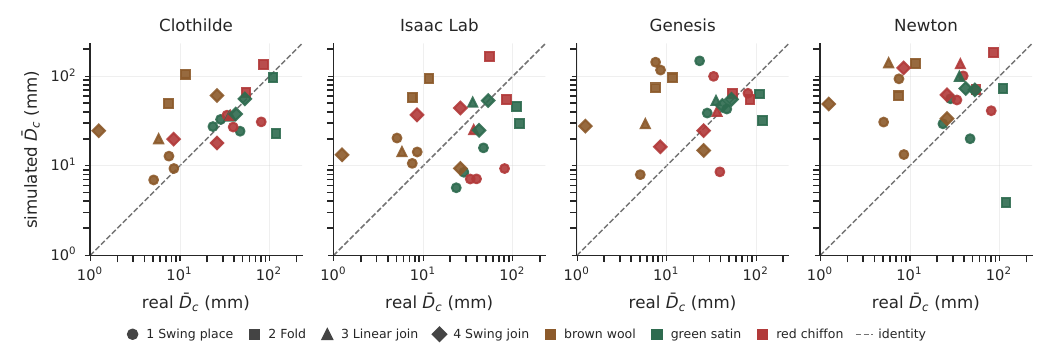}

  \caption{ Comparison of the variability amount in each simulator compared to the OptiTrack ground truth. Points closer to the diagonal correspond to more accurate simulations. None of the simulators captures the real variability.}
  \label{fig:sim}
  \vspace{-3mm}
\end{figure*}

\subsection{Analysis of the variability per task}
\label{sec:res-tasks}

\smallskip\noindent\textbf{Swing place (task~1).}
To visualize the variability of this task, we show the location of the resting points of the bottom edge markers m1 to m5 in \figtaskref{1}. Each dot corresponds to one single marker and rollout execution. 
The darker the blue of the dots, the higher the execution speed. The results clearly show that for the thick cloth, the resting poses of all markers are less than a centimeter apart (5--9\,mm) at every speed. In contrast, the two light cloths scatter 23--47\,mm (green satin) and 33--81\,mm (for the red chiffon), three to eleven times more, and their spread grows with speed while the heavy cloth's hardly does. The ordering by amount of variability follows the fabric: the lighter and the more elastic, the more it disperses and the more speed matters. Every configuration exceeds its floor variability
($\eps_c \le 2.1$\,mm) by 14--1250$\times$.

It's also worth noting that for slower speeds, the variability is mainly in the direction of motion (X axis). However, for light cloths and high speeds, the variability also becomes very noticeable in the Y axis, indicating more complex aerodynamic effects.

\smallskip\noindent\textbf{Dynamic fold (task~2).}
A similar plot is shown for the dynamic fold task in \figtaskref{2}. We show the end location of the marker that swings to fold (m5), and the target location, that is, the position of the marker m15. However, note that we do not care about task success, but on how repeatable the folding motion is.
The free corner lands away from its target in every configuration, by 7 to 41\,cm. Clearly the best behaved is the thick fabric, with a dispersion of 7--12\,mm against 55--118\,mm for the light ones. We should also consider that, for the light cloths, the flap sometimes crumples or folds instead of landing flat, introducing even more error.
Slowing the speed by 40\,\% does not significantly affect the result. 

In this case, the floor variability measure is below 1\,mm, 39--800 times less than at the free corner.
These results indicate that this task is highly sensitive to cloth properties, and an optimized policy that reaches the target is unlikely to generalize across cloth types.

\smallskip\noindent\textbf{Linear and swing inward gripper motion (task~3 and 4).}
A similar analysis of the resting positions of all markers for tasks 3 and 4 is available on the paper website. Due to space restrictions, we show only the analysis of the buckling side for tasks 3 and 4. 
During the inward motion, the grippers compress the top edge, which buckles out of the line m5--m15 with a magnitude $d_{\perp}$ described in Eq.~\eqref{eq:dperp}. Fig. \ref{fig:joins} shows the values $d_{\perp}$ for each rollout configuration. On the top, we see the values for the linear inward motion, and at the bottom the values when the swing is introduced.

First, for the stiff wool fabric, we see the task is quite repeatable, and even without the swing, the cloth was already buckling always to the same side with similar magnitude. The swing introduces more variability in the magnitude of $d_{\perp}$, which in this case may be positive to achieve a fold in the air but is less repeatable. For the green satin, we can see how at the same speed ($\ts = 1$), the swing clearly forces the buckling towards the same side, also for the regrasping case. However, at a slower speed, the dynamics are not enough to ensure the direction of the buckling towards the intended side. Finally, for the very soft red chiffon cloth, the swing does not seem to influence the outcome at all, at any speed. Looking at the videos, it is clear that for the swing to have the intended effect, the grippers should come much closer before the swing finishes; otherwise, the cloth is soft enough to slide back to the other side in between the grippers. Even for humans, the swing motion with such a soft cloth is challenging.

\subsection{Simulators}
\label{sec:sim}
Fig.~\ref{fig:sim} plots the simulated dispersion of every calibrated engine against the real one (see Sec.~\ref{sec:sim-protocol}). For a perfect simulation, we would see all dots on the exact diagonal, sorted by the variability amount. Note that, even for deterministic simulators like Clothilde and Genesis, each rollout produces a different simulation, since each replay uses the measured start pose and gripper motion from its real rollout.

In Fig.~\ref {fig:sim}, we can clearly see that no engine reproduces the real variability. Clothilde and Genesis are the simulators that get more points on the diagonal. Clothilde performs particularly well with the green satin cloth, followed by Genesis. For the red light cloth, Clothilde and Genesis are also the best. However, the brown one seems challenging for all simulators, except for Clothilde on the swing task, which gets it right.
If we project the points on the X axis, we get the ordering by amount of variability in the real cloth, while projecting on the Y axis gets the order by amount of variability in simulation. We can see that the order is not well preserved in any simulator, with Clothilde being slightly better thanks to  the swing to place task for which works very well.
In addition, we can observe that the same cloth under-disperses one task and over-disperses another, so the dispersion is not related to cloth properties in the simulators.

Two of the simulators, Isaac Lab and Newton, are not deterministic, so identical inputs do not replay identically. This noise is often as large as the real variability, but it is large where the real variability is small, and vice versa. In Isaac Lab, it does not change with the cloth or the task; in Newton, it peaks in the brown linear join, the most repeatable real case, and vanishes in the fold, the least. Therefore, the engine noise is not a replacement for the real variability \footnote{Additional simulation analysis appears on the website.}.

\section{Measuring the dispersion without markers}
\label{sec:vision}

Motion capture is the most time-consuming part of the data collection because it often requires manual re-labeling of markers that get lost for some frames. In addition, most deployed policies use RGB(D) cameras; therefore, in this section we investigate how much of the same analysis as in Section \ref{sec:res-cross} we can perform without markers, using data from the stereo camera that recorded all our experiments. To do this, each terminal state image is first segmented with SAM~2~\cite{ravi2024sam2}, followed by 3-D extraction with FoundationStereo~\cite{wen2025foundationstereo}, which is then described by DINOv3~\cite{simeoni2025dinov3} patches. Each patch takes the role of a virtual marker, in the sense that we compute  $D_c(i,j)$ in Eq.~\eqref{eq:Dcij} using the mean distance between the matched patches. 

\begin{figure}[t]
  \centering
  \includegraphics[width=0.7\columnwidth]{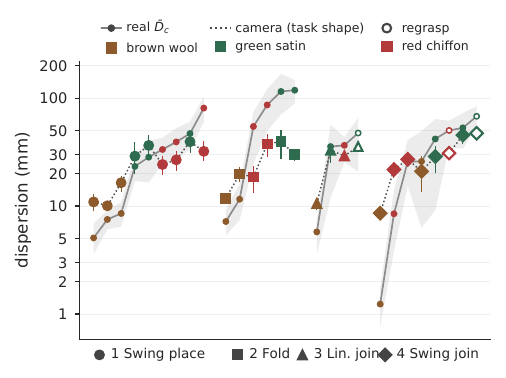}
  \caption{Variability using OptiTrack vs. using the markerless vision estimation. Although the trends are similar, vision does not capture the same variability values, nor the ordering.}
  \label{fig:vision}
  \vspace{-3mm}
\end{figure}

Fig.~\ref{fig:vision} shows the real variability of each task sorted in ascending order, and the corresponding variability using vision. We can see how the general trend in the ordering is preserved, but often variability is overestimated while others is underestimated. In addition, perception struggles most in the dynamic fold task, where the free corner lands on the cloth, and as they have the same color,  the contrast is lost, and it does not define the corresponding patch correctly, leading to big under-estimation of variability.

Therefore, the stereo camera with off-the-shelf models supports a qualitative comparison of cloths, speeds, and tasks, but not a measurement of the dispersion itself with the precision provided by OptiTrack. More details are on the website.

\vspace{-3mm}

\section{Conclusions}

In this paper, we have quantified and characterized the variability in dynamic cloth manipulation in 269 collected executions across four dynamic tasks with a total of 27 different trial configurations.  Of the four tasks in the dataset, two are well known and two are novel tasks. 

Our results show that variability is significant across all experimental configurations, and that it depends highly on the cloth properties, and to a lesser extent on speed. The dramatic change in variability depending on cloth type confirms how important it is to properly report the cloth physical properties in any robotic experiment to really comprehend the capability of any evaluated method.

Moreover, we evaluated this variability in four modern cloth simulators; our findings show that none of them reproduce the variability we observed in real trials. Thus, any sim-to-real approach with dynamic motions will always be a rough approximation unless we can reproduce this behavior in the simulation.
Furthermore, we open-source our labeled dataset of marked cloth keypoints from motion capture, along with the synchronized stereo stream and proprioception, as well as additional analysis on this dataset. We believe this dataset can be useful to train perception methods,  evaluate cloth simulators and learn cloth dynamic models.

In future works, we aim to tackle this sim-to-real gap and use the measured variability to improve the generalizability of trained policies across cloth types. 

\vspace{-1mm}

\bibliographystyle{IEEEtran}
\bibliography{bib/bstctl,bib/IEEEabrv, bib/intro_sota_refs,bib/sim_refs,bib/vision_refs,bib/zotero_references}

\end{document}